# APPROXIMATE DEDUCTION IN SINGLE EVIDENTIAL BODIES[1]

Enrique H. Ruspini
*Artificial Intelligence Center*
*SRI International*
*Menlo Park, CA 94025, U.S.A.*

Abstract:
Results on approximate deduction in the context of the calculus of evidence of Dempster-Shafer and the theory of interval probabilities are reported.

Approximate conditional knowledge about the truth of conditional propositions was assumed available and expressed as sets of possible values (actually numeric intervals) of conditional probabilities. Under different interpretations of this conditional knowledge, several formulas were produced to integrate unconditioned estimates (assumed given as sets of possible values of unconditioned probabilities) with conditional estimates. These formulas are discussed together with the computational characteristics of the methods derived from them.

Of particular importance is one such evidence integration formulation, produced under a belief-oriented interpretation, which incorporates both *modus ponens* and *modus tollens* inferential mechanisms, allows integration of conditioned and unconditioned knowledge without resorting to iterative or sequential approximations, and produces elementary mass distributions as outputs using similar distributions as inputs.

## 1 Introduction

This paper presents results of research on conditionalization issues in the the evidential theory of Dempster-Shafer [1] and the theory of interval probabilities [3,4]. In the context of our work, the term conditionalization is used to denote the processes by which quantitative estimates of the conditional truth of certain propositions (i.e., produced under the assumption of truth of other propositions) are used to refine existing, *a priori*, estimates of unconditioned propositional truth.

In this discussion, the conditioning information is assumed to consist of a number of conditional propositions *(i.e., propositions of the type "if P then Q.")* having associated probability values $p(P \to Q)$ [2] which lie in some subinterval of the $[0, 1]$ interval of the real line. In addition, the scope of this paper is limited to approximate deduction within single evidential bodies. In this context, deduction is the process of combining unconditioned estimates of possible propositional probability values with their conditional counterparts.

In that sense, the problem being considered is different from that leading to results such as the *Dempster's Combination Formula* [2] which results in a consensual quantitative value of propositional truth estimates produced under different, independent, evidential background conditions. In our case, both unconditioned and conditioned estimates are assumed to be valid within the context provided by a single body of evidence. In most practical applications, the unconditioned estimates are the result of analysis of background evidence while conditional estimates (i.e., *approximate knowledge* provided by domain experts) usually have a much wider scope of validity. The

---

[2]Interpreted throughout this paper as being equivalent to the conditional probability $p(P/Q)$.

215

*certainty factors* of MYCIN [5] are examples of conditional truth estimates (albeit in different truth representation and manipulation frameworks).

The major objective of the research being reported was the extension of approximate deduction techniques (i.e. use of *approximate rules*) to frameworks where uncertainty and ignorance are properly differentiated, and where incorporation of additional information within an evidential framework results in a decrease of the ignorance about propositional truth (the decrease being null only when irrelevant conditional information is considered).

In addition, other objectives of this research included:

1. The use of other deductive procedures (e.g., *modus tollens*) in addition to those typically implemented within expert systems with approximate reasoning capabilities (primarily extensions of *modus ponens*).

2. The use of both conditioned and unconditioned estimates as joint *constraints* on the values of valid probability distributions (valuations) on a propositional lattice.

3. The use of proper reference classes [6] to propagate constraints between antecedent and consequent propositions in a conditional propositional statement (i.e., if it is known that $x \to y$ and that $z_1, z_2, \cdots, z_n$ are true should one use $p(x/y)$ or $p(x/y, z_1, z_2, \cdots, z_n)$ to propagate evidence ?).

4. The development of deductive algorithms that properly use constraints defined by interval-valued truth estimates producing results that are not affected by changes in the implicit or explicit inferential control structure .

## 2 Preliminaries

### 2.1 Beliefs and Envelopes

Differences between *probability envelopes* and Shafer's *belief functions* are first discussed as a necessary prerequisite for the understanding of conditionalization results discussed below.

*Probability Envelopes* of a convex family $C$ of probability functions in a propositional lattice $L$ [3] are bounds satisfied by elements of $C$ that cannot be improved without finding members of $C$ which do not satify the improved bounds:

**Definition:** The function $b : L \to [0,1]$ is said to be a *probability envelope* for the set of probability functions $C$ in $L$, if and only if for every $x$ in $L$, there exists a probability function $p_x$ in $C$ such that:

(a) $p_x(x) = b(x)$,

(b) $p_x(y) \geq b(y)$, for all $y$ in $L$.

*Belief functions*, on the other hand, are characterized in terms of mappings between a space of possible worlds and a propositional space [2]. Belief functions are real functions induced in the image propositional space by probability distributions in the domain of the mapping. From the perspective of such a mapping, belief functions generalize the logical characterization of probabilities of Carnap (as additive measures over *spaces of possible worlds*) to logics accepting three values of

---

[3] Throughout this paper, probabilities will be treated either as valuations on a complemented lattice, or alternatively as real functions defined over subsets of some space depending on the particular discussion context.

216

propositional truth (i.e., **true**, **false**, **possible**). It is straightforward to see that belief functions are envelopes.

In what follows, the expression *support function* will be used to denote either belief function or probability envelope whenever the stated results apply to functions in either of those functional classes.

## 2.2 Specifity

The frameworks studied in this paper for the representation of imprecision and uncertainty allow explicit representation of the degree of ignorance about probability values by means of subintervals of the $[0, 1]$ interval of the real line. Different estimates can be partially ordered according to their *specificity*, i.e. their ability to precisely specify the unknown probability values:

**Definition:** If $b_1$ and $b_2$ are support functions in the propositional lattice $L$ then then $b_1$ is said to be more *specific* than $b_2$ if and only if $b_1(x) \geq b_2(x)$, for all $x$ in $L$.

This notion of specificity, due to Yager [7], is important to understand the purpose of any evidence combination procedure, that of reducing uncertainty about the state of systems being modeled or observed. It can be informally said that the purpose of evidence aggregation is the derivation of evidence which is more specific (i.e., provides a more detailed characterization of *possible* states) than each of the evidential bodies being combined.

## 2.3 Conditional Information

The major objective of the research work reported in this paper was the identification and study of algorithms that combine a support function $b$ defined in a lattice $L$ with information on the validity of conditional propositions (i.e., *knowledge* about the probability of $x \to y$, where $x$ and $y$ are elements of $L$) in order to determine a more specific (i.e *improved*) support function $\hat{b}$.

Conditional information is assumed to be given in the form of a function $c$ (called a *conditional evidence* or a *conditional support function*) such that

$$c : L \times L \to [0, 1],$$

and

(a) $c(\cdot, y) : L \to [0, 1]$ is a support function if $y \neq \emptyset$ or $y \neq \Theta$,[4]

(b) $c(x, y) = c(x \wedge y, y)$ if $y \neq \emptyset$.

Two alternative interpretations of the notion of conditional evidence will be discussed. First, in the case of interval probabilities, the functional value $c(x, y)$ may be regarded as a lower bound on the value of the conditional probability $p(x/y)$. Second, in a belief function context, the function $c$ indicates the amount of belief that shall be committed to the truth of a proposition *solely* on the knowledge that another proposition is true.

---

[4]Following customary practice in Dempster-Shafer theory we denote the infimum and the supremum of $L$ by $\emptyset$ and $\Theta$, respectively.

## 2.4 Consistence

Informally, an unconditioned support function $b$ and a conditional evidence function $c$ are *consistent* whenever the information provided by $c$ cannot be used to improve (i.e. make more specific) the space of possible probability values defined by $b$.

In general such consistence will not be present whenever an arbitrary support function $b$ (representing background knowledge on the truth of elementary propositions) is considered together with domain knowledge relating possible values of probabilities for different propositions[5]. The purpose of any procedure to combine conditioned and unconditioned information may therefore be thought of as the production of a more specific support function $\hat{b}$ that is *consistent* with the conditional knowledge $c$.

## 3 Conditional Probability Constraints

### 3.1 Consistence Considerations

If the interpretation of conditional support is restricted to be one of lower bounds on conditional probabilities, then, informally, a support function $b$ and a conditional support function $c$ are *consistent* if the minimum value that can be attained by a conditional probability under the constraints specified by $b$ is larger than the corresponding value specified by $c$.

In order to provide more flexibility to the definition of consistence, it will be assumed that an additional hypothesis set $\mathcal{H}$ [6] has been specified to further constrain the extent of possible probability distributions. With this consideration in mind, consistence between support functions can be defined in the following way:

**Definition:** The support functions $b$ and $c$ are said to be *consistent* under $\mathcal{H}$ if and only if

$$\inf_{\mathcal{H},b}[p(x/y)] \geq c(x,y), \qquad \text{for all } x,y \text{ in } L.$$

Under consistence, therefore, all probability functions that satisfy both the constraints $p(x) \geq b(x)$ for all x in $L$, and the additional hypotheses $\mathcal{H}$, are such that the corresponding conditional probability values $p(x/y)$ satisfy:

$$p(x/y) = \frac{p(x \wedge y)}{p(y)} \geq c(x,y), \qquad x,y \text{ in } L.$$

If $b$ and $c$ are consistent then the unconditioned bounds defined by $b$ cannot be improved as the conditional information does not provide any new additional knowledge. If, on the other hand, $b$ and $c$ are not consistent then the bounds may be improved by modification of $b$ to yield a new support function $\hat{b}$. This improved function is necessarily more specific than $b$ [7].

---

[5] Otherwise, the *conditional knowledge* would add no new information and there would be no purpose to use it in the context of a *knowledge-based system*.

[6] For example, it may be known that certain propositions in $L$ are *independent*.

[7] Clearly any improvement in either $c$ or $b$ will lead to support functions which should be more specific than either of them (as addition of knowledge cannot lead to more ignorance). Improving $c$, however, does not alleviate the inability to meet conditional bounds required by the definition of consistence. This situation can be remedied, however, by making $b$ more specific thus reflecting the conditional information provided by $c$.

### 3.2 Generalization of a Bayesian Identity

If p is a probability defined in the lattice $L$, a well known identity asserts that

$$p(x) = p(x/y)p(y) + p(x/\bar{y})(1 - p(y)), \qquad (1)$$

for all $x$ in $L$, and for all $y$ in $L$ such that $0 < p(y) < 1$.

Now, if $b$ and $c$ are unconditioned and conditional support functions, respectively, it follows at once that any probability $p$ consistent with them will satisfy:

$$p(x) \geq c(x,y)p(y) + c(x,\bar{y})(1 - p(y)),$$

and since $b(y) \leq p(y) \leq 1 - b(\bar{y})$, then

$$p(x) \geq \min_{\alpha}\left[c(x,y)\alpha + c(x,\bar{y})(1-\alpha)\right],$$

where the minimum is taken over real $\alpha$ satisfying $b(y) \leq \alpha \leq 1 - b(\bar{y})$. Since this inequality must be satisfied for all $p$ consistent with $b$ and $c$ it makes sense to require that

$$b(x) \geq \min_{\alpha}\left[c(x,y)\alpha + c(x,\bar{y})(1-\alpha)\right], \qquad (2)$$

where $b(y) \leq \alpha \leq 1 - b(\bar{y})$.

Equation (2) is therefore a generalization of the identity (1) which is of considerable importance in approximate deduction since support functions $b$ and $c$ that satisfy equation (2) can be seen to retain, at a probabilistic level, the important relationships between truth values of antecedent and consequent propositions in implications in two-valued logic, which is the basis for the inferential processes of *modus ponens* and *modus tollens*.

### 3.3 Optimistic Case

In studying possible interpretations to be made of the relationship between consistent support functions $b$ and $c$ it is interesting to consider only certain probability distributions that satisfy additional hypotheses $\mathcal{H}$ which restrict the number of possible probability functions $b$ that may need to be considered.

One such interesting sets of hypothesis $\mathcal{H}$ is that that regards knowledge of the certainty of a proposition $y$ as requiring that the unconditioned bounds $b$ be modified so that

$$\tilde{b}(\bar{y}) = 0, \quad \tilde{b}(x \wedge y) = b(x \vee \bar{y}) - b(\bar{y}),$$

where $\tilde{b}$ indicates the refined bounds. Such an assumption, is equivalent, in the case of belief functions, to the replacement (under the assumption of truth of a proposition $y$) of the mapping $\Gamma$ by the refined mapping $\Gamma \wedge y$. In other words, if a state which was compatible with a proposition $x$ has a possibility of being compatible with $y$ (i.e., $x \wedge y \neq \emptyset$) then it is compatible with $x \wedge y$ (i.e., we ignore the possibility that the true state may be in $x \wedge \bar{y}$).

Under such an *optimistic* assumption, support functions $b$ and $c$ will be consistent if and only if

$$\frac{b(x \vee \bar{y}) - b(\bar{y})}{1 - b(\bar{y})} \geq c(x, y), \qquad (3)$$

over all $x, y$ in $L$ such that $b(\bar{y}) < 1$.

219

It can be further seen that in this case,

$$b(z) \geq \max_{y \geq \bar{z}} \left[ b(\bar{y})(1 - c(z,y)) + c(z,y) \right].$$

This equation is also the basis for a fast iterative method to derive (from a given $b$) the least specific $\hat{b}$ which is more specific than $b$ and consistent with $c$ (under the optimistic assumptions).

Although the optimistic assumption is of particular interest since the expression in the left hand side of the equation (3) is that obtained when using the Dempster combination formula to combine the given $b$ with a simple support function focused in $y$, consistent functions under that assumption may, however, fail to satisfy the equation (2) which assures proper generalization of the relationships between truth values of antecedent and consequent in two valued logic.

### 3.4 General Case

If no additional assumptions are made (i.e. $\mathcal{Y}$ is empty) then it can be seen that consistence, in this most general and unrestricted case, is equivalent to the inequalities

$$\frac{b(x \wedge y)}{1 + b(x \wedge y) - b(x \vee \bar{y})} \geq c(x,y), \tag{4}$$

for all $x, y$ in $L$ such that $b(\bar{y}) < 1$.

It can also be seen that the left hand side of the Equation (3) is always larger than the left hand side of the above equation indicating, as expected, that it is easier for two support functions $b$ and $c$ to be consistent under the optimistic assumption rather than in the general case where the left hand side of equation (4) is the lowest possible value that may be taken by $p(x/y)$ if $p$ is bounded by below by $b$. Further, support functions that satisfy (4) satisfy also the extensions to interval probabilities (equation (2)) of the Bayes identity (1), thus guaranteeing the proper relationships between $b(x), b(y)$ and $c(x,y)$.

## 4 Belief-Based Interpretations

The last part of this paper deals with alternative interpretations of the notion of conditional support which apply solely to unconditioned and conditional support functions which are belief functions.

Under these interpretations, the conditioned support function $c$ is considered to be a family of *quantitative rules* for the redistribution of ignorance (represented by the multivalued mapping $\Gamma$ that underlies Dempster's theoretical formalism). The basic idea is that that the elementary conditional mass functions $m_y(x)$ associated with the conditional supports $c(x,y)$ provide quantitative mechanisms to redistribute the mass associated with the sets $\{s \in S : \Gamma(s) = y\}$ and $\{s \in S : \Gamma(s) \leq y\}$, by redefinition of the mapping $\Gamma$.

These interpretations result in the derivation of straightforward, non-iterative formulas for the production of more specific, consistent, unconditioned supports $\hat{b}$. Further, under one of the interpretations discussed below, the resulting improved support function $\hat{b}$ is also a belief function thus assuring that results of the conditionalization process remain within the theoretical framework of the Dempster-Shafer theory.

### 4.1 Conditional Mappings

Consistent with Dempster's theoretical formalism we assume the existence of a probability space $\{S, \Omega, P\}$, where, for simplicity, it will be assumed that $S$ is finite and that $\Omega$ is its power set. It is also assumed that there exists a nonvoid set T and a multivalued mapping $\Gamma : S \to T$. From these structures, it is possible to define *belief* and *basic probability* or *mass* functions:

$$b(x) = P(\{s \in S : \Gamma(s) \subseteq x\}),$$
$$m(x) = P(\{s \in S : \Gamma(s) = x\}),$$

where $x$ is a subset of $T$.

Extending the basic Dempster-Shafer formalism, we shall also assume that there exists an ordinary mapping

$$\tau : S \to T,$$

such that $\tau(s) \in \Gamma(s)$. The mapping $\tau$, called the *truth function* is used in our formalism to represent the true proposition (represented by a point of $T$) associated with each possible state of the world $s$ in $S$ [8]. It is also necessary to introduce the *truth sets* $T_y = \{s \in S : \tau(s) \in y\}$, where $y$ is a nonvoid proper subset of $T$.

It will be then assumed that there exist for each such $y$ there exists a multivalued mapping:

$$\Gamma_y : T_y \to T,$$

such that the following set identity is true for all nonvoid proper subsets $u$ and $v$ of $T$ such that $u \subseteq v$:

$$\{s \in S : \Gamma_v(s) = x \text{ and } s \in T_v\} = \{s \in S : \Gamma_u(s) = x \cap u \text{ and } s \in T_u\},$$

whenever $x \subseteq v$ [9].

### 4.2 Conditional Beliefs as Simple Bounds

If now $c(.,y) : y \to [0,1]$ is defined as the belief function (in $y \subseteq T$) associated with $\Gamma_y$ [10] and it is further required that

$$P(T_x) \geq c(x, y) P(T_y),$$

for all proper nonvoid $y \subseteq T$, then it can be seen that the most specific probability *lower bound* consistent with $c$ which improves $b$ is given by:

$$\hat{b}(x) = \sup_{P(T)} [\sum_{u \in P} b(u) c(x, u) - \inf_{v \in P} [c(v, x) \rho(P)]],$$

where the supremum is taken over all partitions P of T, and where

$$\rho(P) = 1 - \sum_{v \in P} b(v).$$

---

[8] This function is, of course, unknown. The mapping $\Gamma$ is the modeling tool used to represent our actual knowledge of *possible* propositions consistent with $s$.

[9] The rationale for this equality demanding some *consistence* between *conditional mappings* falls outside the scope of the limited discussion of this paper.

[10] This function may be extended to all of $T$ by means of the equality $c(x, y) = c(x \cap y, y)$.

221

### 4.3 Mass Based Interpretation

If, on the other hand, we interpret the conditional belief functions $c(.,y)$ in terms of their associated masses $m_y(x)$ as constraining the conditional mappings $\Gamma_y$ and the *truth sets* $T_y$ so that:

$$m_y(x) = P(\Gamma_y(s) = x \cap y / T_y),$$

for $x \subseteq T$, then the most specific *belief* consistent with the unconditioned belief $b$ (and its associated mass distribution $m$), is given by:

$$\hat{m}(x) = m(x) \; + \\ + \sum_{y>x} m(y)(\sup_{P(y)}[\inf_{u \in P}(m_u(x))]) - \\ - m(x) \sum_{z<x} \sup_{P(z)}[\inf_{u \in P}(m_u(z))],$$

where the $P(u)$ indicates all partitions of the subset $u$ of $T$.

Unlike the case with the previous interpretation, use of this formula results in a procedure that is guaranteed to produce a belief function as an output. This output is produced in a single pass of the above formula (i.e., not through an iterative procedure). In addition this formula leads to methods that produce mass distributions using as inputs mass distributions thus avoiding the computational complexity problems associated with direct manipulation of belief functions. These efficient computation mechanisms also incorporate (as was the case with all other expressions above) both *modus ponens* and *modus tollens* mechanisms for the derivation of improved estimates of possible truth values.

## 5 Acknowledgments

The author benefitted from discussions and comments by T. Cronin, T. Garvey, S. Lesh, J. Lowrance, T. Strat, L. Wesley, and L. Zadeh. To all of them many thanks.